\documentclass[journal]{IEEEtran}
\IEEEoverridecommandlockouts
\usepackage[T1]{fontenc}
\usepackage{graphicx}
\usepackage{amsmath}
\usepackage[caption=false,font=footnotesize]{subfig}
\usepackage{multirow}
\usepackage{rotating}
\usepackage{dblfloatfix}
\usepackage{float}
\usepackage{xcolor}
\usepackage[hyphens]{url}

\def\BibTeX{{\rm B\kern-.05em{\sc i\kern-.025em b}\kern-.08em
    T\kern-.1667em\lower.7ex\hbox{E}\kern-.125emX}}

\begin{document}

\title{Pseudo-Label Foreground Quality Weighting for Domain-Generalized Fundus Image Segmentation}

\author{%
    \IEEEauthorblockN{Lucas Gabriel Telesco\textsuperscript{1,2,3}, José Ignacio Orlando\textsuperscript{1,2,3}}
    \\
    \IEEEauthorblockA{\textsuperscript{1} Yatiris Group, PLADEMA, UNICEN, Tandil, Argentina}\\
    \IEEEauthorblockA{\textsuperscript{2} CONICET, Tandil, Argentina}\\
    \IEEEauthorblockA{\textsuperscript{3} Retinar S.A.S., Campus Universitario, Tandil, Argentina}\\
    \IEEEauthorblockA{\{ltelesco,jiorlando\}@pladema.exa.unicen.edu.ar}
}

\maketitle

\begin{abstract}
Optic disc (OD) and optic cup (OC) segmentation is central to automated glaucoma assessment, but models trained on limited labeled fundus datasets often degrade under domain shift. Semi-supervised Teacher-Student learning can improve generalization by leveraging unlabeled data, yet standard Noisy Student training treats all pseudo-labels as equally reliable. In this paper we investigate a targeted extension of Noisy Student training in which OD- and OC-specific pseudo-label supervision is modulated using a mask-derived reliability estimate. A multitask Dice predictor estimates OD- and OC-specific pseudo-label quality from Teacher-generated masks, and these estimates are used to weight the pseudo-label loss during Student training. This preserves the baseline contribution of low-scoring foreground pseudo-labels while assigning greater relative emphasis to pseudo-labels predicted to be more reliable. Experiments across known and unseen domains show improvements over the Noisy Student baseline, particularly for OC segmentation, while maintaining competitive generalization performance. %These results support pseudo-label quality estimation as a simple and practical strategy for robust semi-supervised OD/OC segmentation under domain shift.
\end{abstract}


\section{Introduction}

Deep learning supports many fundus-image analysis tasks, including glaucoma screening, disease grading, and anatomical structure segmentation \cite{fu2018mnet,fu2018denet,orlando2020refuge}. For glaucoma assessment, segmenting the optic disc (OD) and optic cup (OC) is especially important as both structures are used to estimate the cup-to-disc ratio, a clinical biomarker of glaucomatous optic neuropathy \cite{fu2018mnet,orlando2020refuge}. Yet models that perform well on curated benchmarks often degrade on unseen data \cite{wang2019posal,wang2019beal,liu2019cfea,wang2020dofe}. Differences in camera vendor, field of view, resolution, retinal pigmentation, pathology, acquisition protocol, and image quality create domain shifts that are underrepresented during training, limiting their generalization and clinical reliability.

\begin{figure}[t]
    \centering
    \includegraphics[width=0.9\columnwidth]{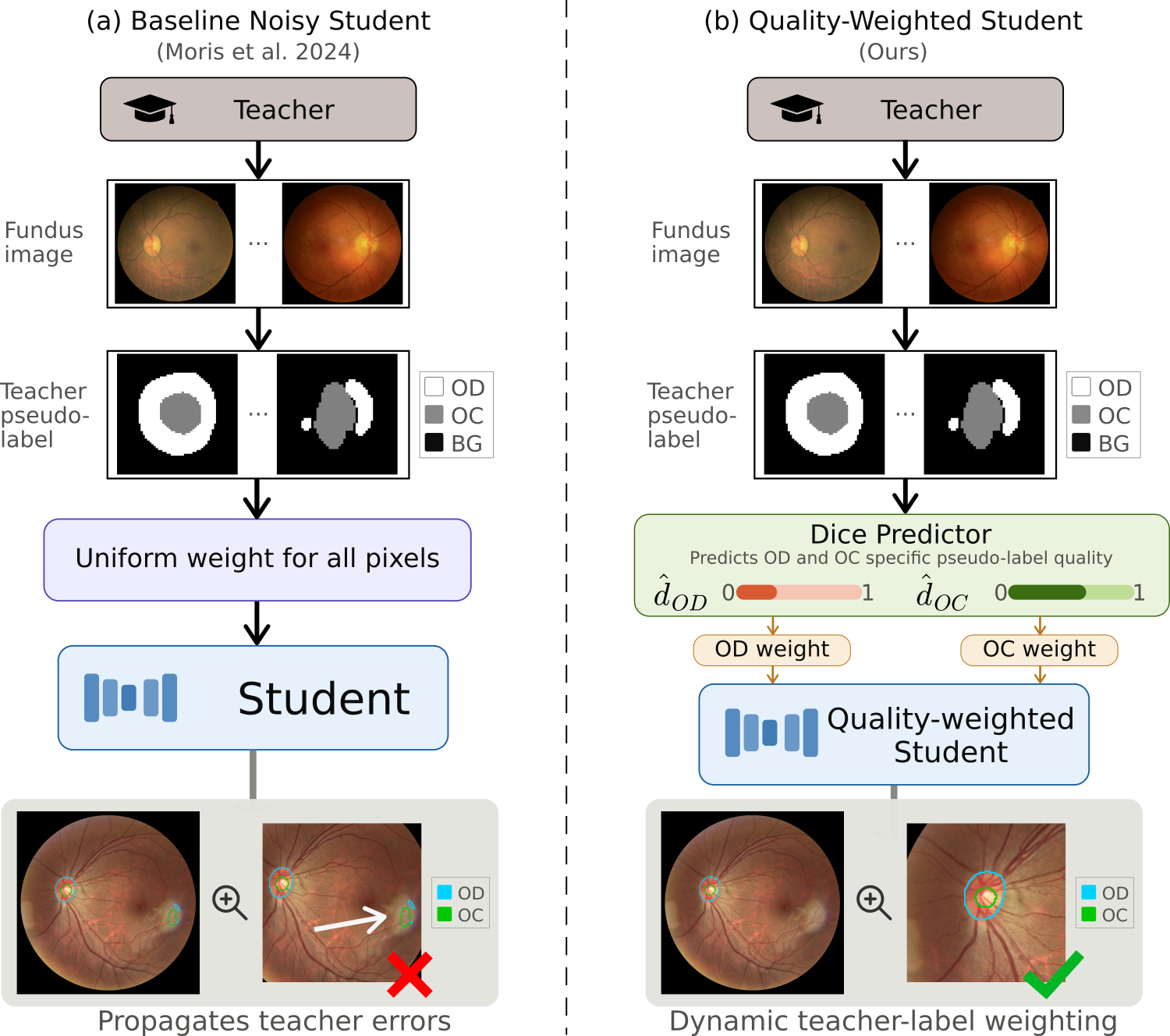}
    \caption{Summary of the proposed approach. (a) Noisy Student method as in \cite{moris2024noisystudent}. (b) Our proposed quality-weighted student.}
    \label{fig:teaser}
\end{figure}

Several methods were recently introduced to address this issue. pOSAL \cite{wang2019posal}, BEAL \cite{wang2019beal}, and CFEA \cite{liu2019cfea}, for example, align labeled source and unlabeled target domains through adversarial learning, output-space alignment, or collaborative ensembling, while DoFE \cite{wang2020dofe} learns domain-oriented features for unseen domains. However, all these were developed on images cropped around the optic nerve. Cropping simplifies segmentation but removes real-world variation in field of view, illumination, peripheral artifacts, and optic-disc location. Moris et al.~\cite{moris2024noisystudent} addressed this using uncropped fundus images and a semi-supervised Noisy Student strategy~\cite{xie2020self}. A Teacher trained on several labeled datasets generates pseudo-labels for a large unlabeled collection, used then to train a Student. This broadens image variability but treats all pseudo-labels as equally reliable, allowing Teacher errors to propagate. In this paper we extend the approach by weighting OD- and OC-specific pseudo-label supervision using a mask-based reliability estimate (Fig.~\ref{fig:teaser}). A quality predictor estimates the expected Dice score of each Teacher segmentation, so more reliable foreground pseudo-labels receive greater weight during Student training. Across known and unseen domains, this strategy improves over the Noisy Student baseline and achieves competitive or better generalization without extra manual annotations or inference-time cost. These results show that structure-specific quality estimates can improve semi-supervised OD/OC segmentation under cross-domain variation. Results are publicly released for comparison.\footnote{\url{https://github.com/ltelesco/pseudo-label-quality-weighting-for-domain-generalized-optic-disc-and-cup-segmentation}}

\section{Methods}

\subsection{Noisy Student for Domain Generalization}

Let $\mathcal{D}_{L}=\{(x_i,y_i)\}_{i=1}^{N}$ be a labeled set of uncropped fundus images with manual OD/OC labels, and let $\mathcal{D}_{U}=\{u_j\}_{j=1}^{M}$ be an unlabeled set collected from heterogeneous imaging conditions. We build on the Noisy Student strategy used by Moris et al.~\cite{moris2024noisystudent}, where a Teacher model $T$ trained on $\mathcal{D}_{L}$ generates hard pseudo-labels $\tilde{y}_j=T(u_j)$ for all unlabeled images, yielding $\tilde{\mathcal{D}}_{U}=\{(u_j,\tilde{y}_j)\}_{j=1}^{M}$. A Student model $S_{\theta}$ is then trained from both manual annotations and pseudo-labels. Being $\ell(\cdot,\cdot)$ the segmentation loss, the baseline objective is:
\begin{equation}
\mathcal{L}_{\mathrm{NS}}(\theta)=\frac{1}{N}\sum_{i=1}^{N}\ell(S_{\theta}(x_i),y_i)+\lambda\frac{1}{M}\sum_{j=1}^{M}\ell(S_{\theta}(u_j),\tilde{y}_j),
\label{eq:baseline-loss}
\end{equation}
where $\lambda$ controls the contribution of pseudo-labeled data.

This formulation improves domain generalization by exposing the Student to a broader image distribution than the labeled training set alone. However, it also treats all pseudo-labels as equally informative, when, in practice, Teacher predictions may vary substantially in quality due to the same domain shifts we are trying to overcome. As a result, inaccurate pseudo-labels may contribute to training with the same strength as reliable ones, making the Student to reproduce Teacher errors.

\subsection{Quality-Weighted Pseudo-Label Supervision}

We make the pseudo-label term quality-aware by associating each Teacher prediction with an estimated segmentation quality. For each unlabeled image $u_j$, a Dice prediction model $Q_{\phi}$ estimates the expected Dice of the Teacher-generated foreground masks as $Q_{\phi}(\tilde{y}_j)=(\hat{d}^{\mathrm{OD}}_j,\hat{d}^{\mathrm{OC}}_j)$, with $\hat{d}^{\mathrm{OD}}_j,\hat{d}^{\mathrm{OC}}_j\in[0,1]$. We denote by $\ell_{\mathrm{BG}}$, $\ell_{\mathrm{OD}}$, and $\ell_{\mathrm{OC}}$ the class-wise contributions of the segmentation loss associated with pixels pseudo-labeled as background, OD, and OC, respectively. The proposed quality-weighted Student objective is
{
\begin{equation}
\begin{array}{rl}
    \mathcal{L}_{\mathrm{QNS}}(\theta)
    =&
    \frac{1}{N}\sum_{i=1}^{N}\ell(S_{\theta}(x_i),y_i) \\
    &+ \lambda \frac{1}{M}\sum_{j=1}^{M}
    \Big[
    \ell_{\mathrm{BG}}(S_{\theta}(u_j),\tilde{y}_j) \\
    &+
    (1+\hat{d}^{\mathrm{OD}}_j)\ell_{\mathrm{OD}}(S_{\theta}(u_j),\tilde{y}_j) \\
    &+
    (1+\hat{d}^{\mathrm{OC}}_j)\ell_{\mathrm{OC}}(S_{\theta}(u_j),\tilde{y}_j)
    \Big].
\end{array}
\label{eq:quality_weighted_objective}
\end{equation}
}
Thus, the supervised term remains unchanged, and only the pseudo-label contribution is modified. OD and OC pixels are weighted by the predicted quality of their foreground pseudo-label: since predicted Dice scores lie in $[0,1]$, foreground weights range from $1$ to $2$, giving reliable pseudo-labels greater influence while keeping uncertain ones close to the baseline background. Moreover, notice weighting is structure-specific, so a pseudo-label can still provide useful OD supervision even when its OC prediction is unreliable, or vice versa.

\subsection{Training the Dice Prediction Model}

The Dice prediction network $Q_{\phi}$ is trained as a multitask regression model using Teacher predictions for labeled images, where their true quality can be measured against manual annotations. Let $\mathcal{D}_{Q}\subseteq\mathcal{D}_{L}$ denote the labeled set used for this stage. For each $(x_i,y_i)\in\mathcal{D}_{Q}$, we obtain the Teacher prediction $\bar{y}_i=T(x_i)$ and compute the classical Dice coefficient $d_i^c=2|\bar{y}_i^c\cap y_i^c|/(|\bar{y}_i^c|+|y_i^c|)$ for each foreground structure $c\in\{\mathrm{OD},\mathrm{OC}\}$, where $\bar{y}_i^c$ and $y_i^c$ are the corresponding binary masks. These Dice coefficients are used only as regression targets for quality estimation, not as a segmentation loss. Then, the quality predictor receives the Teacher-generated segmentation as input and outputs $Q_{\phi}(\bar{y}_i)=(\hat{d}_i^{\mathrm{OD}},\hat{d}_i^{\mathrm{OC}})$, with OD and OC scores predicted by two separate fully connected heads. The model is trained by minimizing:
\begin{equation}
    \mathcal{L}_{Q}(\phi)
    =
    \frac{1}{|\mathcal{D}_{Q}|}
    \sum_{(x_i,y_i)\in\mathcal{D}_{Q}}
    \left(
    |\hat{d}_i^{\mathrm{OD}}-d_i^{\mathrm{OD}}|
    +
    |\hat{d}_i^{\mathrm{OC}}-d_i^{\mathrm{OC}}|
    \right).
    \label{eq:dice_predictor_loss}
\end{equation}
After training, $Q_{\phi}$ is frozen and applied to the pseudo-labels generated for $\mathcal{D}_{U}$. The predicted Dice scores are used as fixed reliability weights during Student training. Therefore, the Dice predictor is required only during training and introduces no additional computational cost at inference time.

\begin{figure*}[t!]
    \centering
    \subfloat[]{%
        \includegraphics[width=0.49\textwidth]{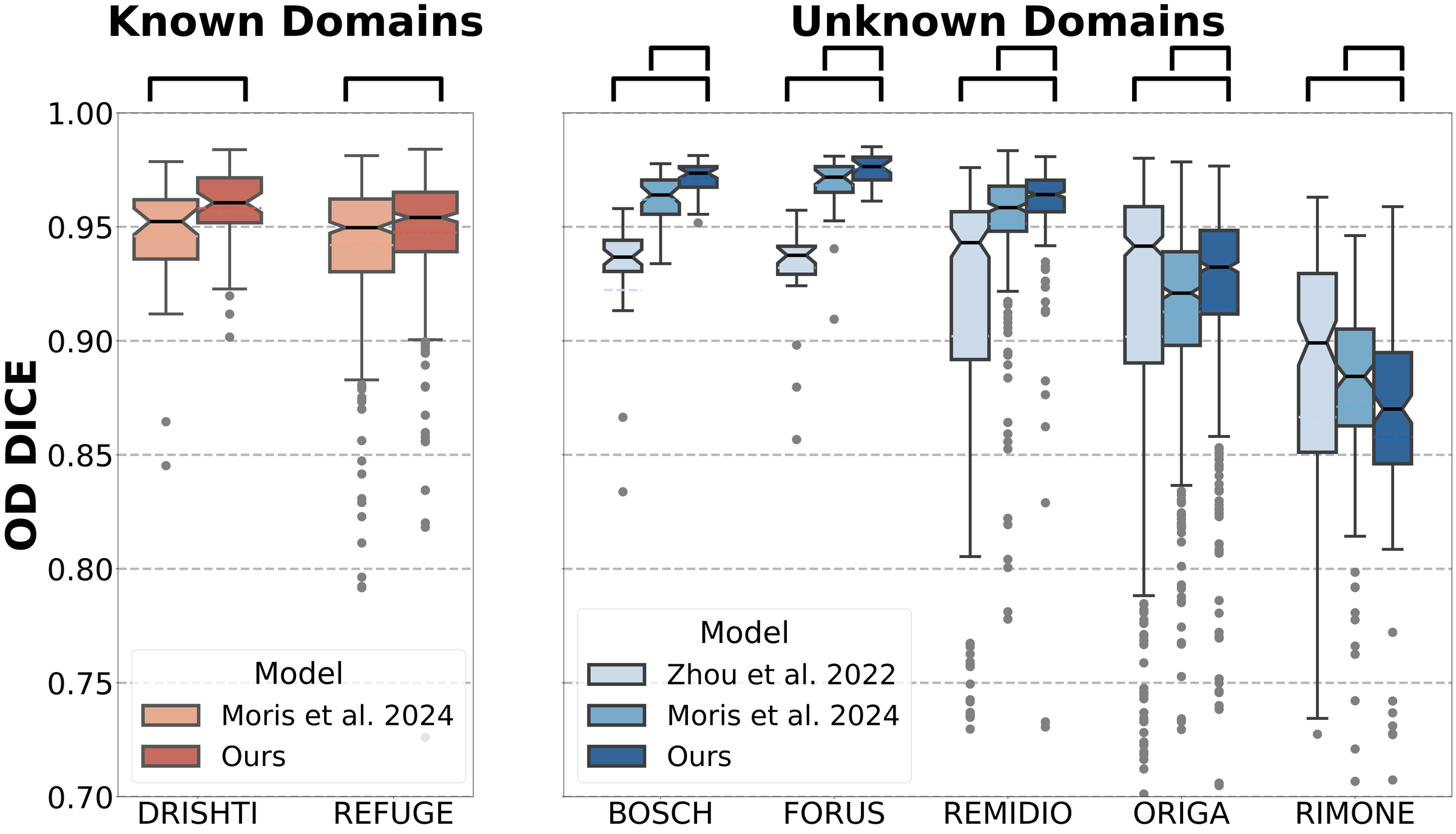}%
        \label{fig:imagen_a}%
    }
    \hfill
    \subfloat[]{%
        \includegraphics[width=0.49\textwidth]{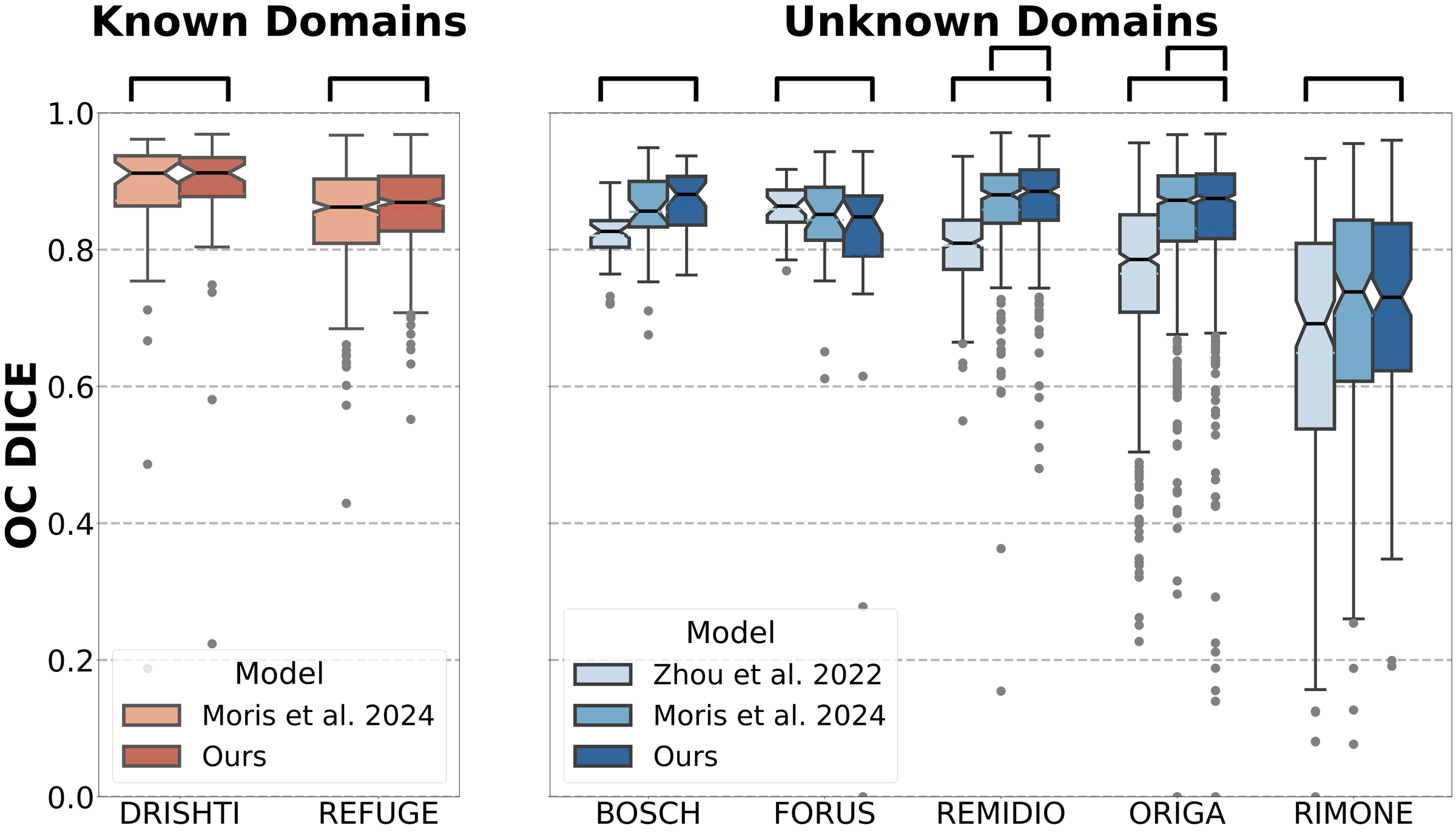}%
        \label{fig:imagen_b}%
    }
    \caption{Dice score distributions for (a) optic disc (OD) and (b) optic cup (OC) segmentation across known (left) and unknown domains (right). \textcolor{black}{In the legends, Moris et al. refers to~\cite{moris2024noisystudent} and Zhou et al. to~\cite{zhou2022generalizable}.} Brackets on top indicate significant differences ($p<0.05$, Wilcoxon signed-rank test).}
    \label{fig:box-plot-dice}
\end{figure*}

\section{Experimental Setup}

For comparability with~\cite{moris2024noisystudent}, we used the same data splits. Labeled images were drawn from DRISHTI~\cite{sivaswamy2014drishti}, REFUGE~\cite{orlando2020refuge}, and RIGA~\cite{almazroa2018riga}, with 45/5, 400/400, and 675/74 images for supervised training/validation, respectively. The unlabeled set $\mathcal{D}_{U}$ was the same AIROGS~\cite{devente2023airogs} subset, with $16{,}200$ training and $1{,}800$ validation images. Known-domain evaluation used the DRISHTI and REFUGE test sets (50 and 400 images), while unknown-domain evaluation used RIM-ONE V3~\cite{fumero2011rimone} (151), ORIGA~\cite{zhang2010origa} (647), and the Ch\'aksu subsets~\cite{kumar2023chaksu}: BOSCH (41), FORUS (31), and REMIDIO (264).

Teacher and Student models followed the configuration in~\cite{moris2024noisystudent}: a three-class U-Net trained on $256\times256$ images to segment background, OD, and OC, using mixed mini-batches of 20 images, with 2 labeled and 18 pseudo-labeled samples, and the same data augmentation strategy. The Dice predictor $Q_{\phi}$, on the other hand, was implemented as a multitask ResNet18~\cite{he2016deep} pre-trained on ImageNet. It receives Teacher-generated $256\times256$ segmentation masks encoded as class-index maps and replicated across three channels, and two independent regression heads estimate OD and OC Dice scores. Training was performed for 200 epochs with Adam optimization (learning rate $0.001$, batch size $64$), using a MultiStepLR schedule with a reducing factor of $0.5$ at epochs 50, 100, and 150. The checkpoint with the lowest validation MAE averaged over OD and OC was kept. To this end we used OD/OC mask pairs from DRISHTI, REFUGE, and RIGA (1{,}118/481 for training/validation), using Teacher predictions on these images as inputs and their Dice as targets.  

Results were compared against~\cite{moris2024noisystudent} and~\cite{zhou2022generalizable} (the best competitor model according to~\cite{moris2024noisystudent}) using Dice, both in known and unknown domains. \textcolor{black}{We also included results of~\cite{moris2024noisystudent} trained on subsets of AIROGS with predicted Dice results above thresholds $\tau \in \{0.1, 0.3\}$, as it is a natural baseline between the original Noisy Student formulation and out proposed approach}. $Q_{\phi}$ results were evaluated against EvanySeg~\cite{senbi2025evanyseg} variants using OD/OC MAE between predicted and actual Dice (significance was assessed with paired Wilcoxon signed-rank tests at $\alpha=0.05$), and Pearson $r$.

\section{Results}

Fig.~\ref{fig:box-plot-dice} compares our proposed approach with the baselines. Table~\ref{tab:dice-mean-std} complements the figure with additional results for the Noisy Student after filtering AIROGS images with different thresholds $\tau$ for predicted Dice. On known domains, our model significantly improves both segmentations in DRISHTI and REFUGE ($p<2.30\times10^{-2}$). On unknown domains, it significantly improves OC results over \cite{moris2024noisystudent} on REMIDIO and ORIGA, and~\cite{zhou2022generalizable} on all evaluated datasets ($p<1.10\times10^{-3}$). For OD, it outperforms both baselines on BOSCH, FORUS, and REMIDIO ($p<1.05\times10^{-5}$), while~\cite{zhou2022generalizable} performs significantly better on ORIGA and RIMONE ($p<1.01\times10^{-4}$).

\begin{table*}[t]
\color{black}
\centering
\caption{\color{black}Dice score (\%) mean $\pm$ standard deviation for optic cup (OC, top) and optic disc (OD, bottom) segmentation across known and unknown domains. Best result per column is highlighted in bold. $^*$ indicates a statistically significant difference against Ours.}
\label{tab:dice-mean-std}
\renewcommand{\arraystretch}{1.12}
\setlength{\tabcolsep}{3pt}
\resizebox{0.9\textwidth}{!}{%
\color{black}
\begin{tabular}{c|c|cc|ccccc}
\hline
& \multirow{2}{*}{\textbf{Model}} & \multicolumn{2}{c|}{\textbf{Known domains}} & \multicolumn{5}{c}{\textbf{Unknown domains}} \\ \cline{3-9}
& & \textbf{DRISHTI} & \textbf{REFUGE} & \textbf{BOSCH} & \textbf{FORUS} & \textbf{REMIDIO} & \textbf{ORIGA} & \textbf{RIMONE} \\ \hline
\multirow{5}{*}{\rotatebox{90}{\textbf{OC}}}
& Zhou et al.~\cite{zhou2022generalizable} & -- & -- & 82.05 $\pm$ 4.08$^*$ & \textbf{85.90 $\pm$ 3.72}$^*$ & 80.58 $\pm$ 6.14$^*$ & 76.48 $\pm$ 12.02$^*$ & 64.89 $\pm$ 20.99$^*$ \\ \cline{2-9}
& Moris et al.~\cite{moris2024noisystudent} & 87.15 $\pm$ 13.10$^*$ & 85.08 $\pm$ 7.21$^*$ & 85.52 $\pm$ 5.94 & 84.34 $\pm$ 7.31 & 85.83 $\pm$ 8.87$^*$ & \textbf{83.09 $\pm$ 14.52}$^*$ & 70.37 $\pm$ 18.23 \\ \cline{2-9}
& + filtered set ($\tau=0.1$) & 86.83 $\pm$ 13.14 & 84.60 $\pm$ 10.18 & 88.22 $\pm$ 4.03 & 76.36 $\pm$ 27.56 & 84.56 $\pm$ 14.99 & 70.63 $\pm$ 30.99 & 68.54 $\pm$ 17.04 \\ \cline{2-9}
& + filtered set ($\tau=0.3$) & 86.78 $\pm$ 12.27 & 85.84 $\pm$ 7.51 & \textbf{89.76 $\pm$ 3.16} & 73.58 $\pm$ 30.50 & 86.25 $\pm$ 8.43 & 82.45 $\pm$ 14.65 & 65.79 $\pm$ 22.91 \\ \cline{2-9}
& Ours & \textbf{88.16 $\pm$ 11.65} & \textbf{86.09 $\pm$ 6.19} & 86.85 $\pm$ 4.72 & 79.31 $\pm$ 19.07 & \textbf{86.85 $\pm$ 7.43} & 82.84 $\pm$ 16.59 & \textbf{70.85 $\pm$ 16.81} \\ \hline\hline
\multirow{5}{*}{\rotatebox{90}{\textbf{OD}}}
& Zhou et al.~\cite{zhou2022generalizable} & -- & -- & 92.22 $\pm$ 7.60$^*$ & 93.20 $\pm$ 2.02$^*$ & 90.20 $\pm$ 9.64$^*$ & 90.18 $\pm$ 9.97$^*$ & 86.65 $\pm$ 10.14$^*$ \\ \cline{2-9}
& Moris et al.~\cite{moris2024noisystudent} & 94.59 $\pm$ 2.49$^*$ & 94.20 $\pm$ 3.04$^*$ & 96.19 $\pm$ 1.05$^*$ & 96.82 $\pm$ 1.40$^*$ & 95.13 $\pm$ 3.12$^*$ & 91.27 $\pm$ 4.16$^*$ & \textbf{87.11 $\pm$ 7.04}$^*$ \\ \cline{2-9}
& + filtered set ($\tau=0.1$) & 95.17 $\pm$ 2.30 & 94.04 $\pm$ 3.00 & 96.79 $\pm$ 0.78 & 97.01 $\pm$ 1.68 & 95.53 $\pm$ 4.44 & 90.37 $\pm$ 12.97 & 86.14 $\pm$ 9.48 \\ \cline{2-9}
& + filtered set ($\tau=0.3$) & 94.58 $\pm$ 2.27 & 93.79 $\pm$ 3.04 & 96.39 $\pm$ 1.03 & 97.25 $\pm$ 1.77 & 95.45 $\pm$ 4.42 & 91.27 $\pm$ 7.24 & 81.25 $\pm$ 21.19 \\ \cline{2-9}
& Ours & \textbf{95.82 $\pm$ 1.95} & \textbf{94.76 $\pm$ 2.79} & \textbf{97.12 $\pm$ 0.69} & \textbf{97.52 $\pm$ 0.65} & \textbf{95.80 $\pm$ 3.39} & \textbf{91.52 $\pm$ 8.42} & 85.77 $\pm$ 8.33 \\ \hline
\end{tabular}%
}
\end{table*}

\begin{figure}[t]
    \centering
    \includegraphics[width=0.95\columnwidth]{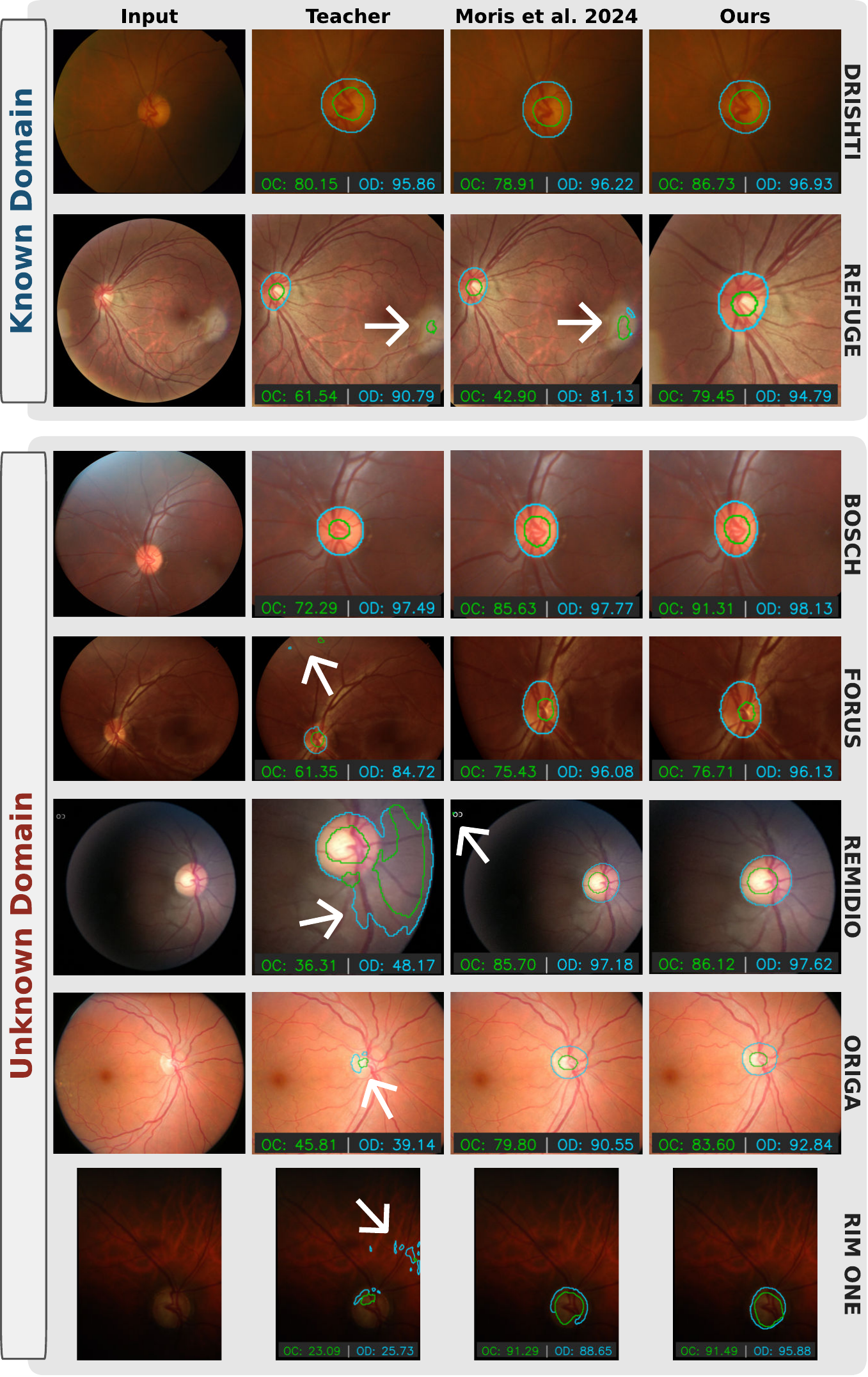}
    \caption{Qualitative comparison of OD (light blue) and OC (light green) results obtained with the Teacher, Moris \textit{et al.}~\cite{moris2024noisystudent} and our approach. Each prediction includes below per-image Dice scores for OC and OD (as percentages).}
    \label{fig:qualitative-comparison}
\end{figure}

Fig.~\ref{fig:qualitative-comparison} shows qualitative examples of improvements thanks to using quality weighting during training. On DRISHTI, for example, the weighting enabled to obtain a better OC segmentation without affecting the quality of the OD. In REFUGE, on the other hand, the baseline exacerbated the mistakes commited by the Teacher in the illumination artifact (see white arrow), while the proposed approach eliminated that mistake and achieved a more consistent segmentation. In unknown domains, the model is improving over the baseline as well, even in scenarios in which the competitive approach was performing already quite well, while fixing tiny mistakes such as false positive pixels within the small text in the image corner.
On ORIGA, for instance, the Teacher only partially segmented the OD and OC regions, the baseline Student~\cite{moris2024noisystudent} already achieved a much better segmentation, and our approach further refined it. On RIM-ONE, however, the Teacher segmented a region outside the area of interest; the baseline corrected this error but still produced a partial OD segmentation, while our approach completed the OD segmentation and also slightly improved OC.

\begin{table*}[t]
\centering
\caption{Dice prediction results for OC (top) and OD (bottom) obtained using~\cite{senbi2025evanyseg} and our multitask network, as evaluated in terms of MAE. Best results are highlighted in bold. $^*$ indicates a statistically significant difference against $Q_{\phi}$.}
\label{tab:dice-predictor-results}
\renewcommand{\arraystretch}{1.12}
\setlength{\tabcolsep}{3pt}
\resizebox{0.8\textwidth}{!}{%
\begin{tabular}{c|c|c|cc|ccccc}
\hline
& \multirow{2}{*}{\textbf{Model}} & \multirow{2}{*}{\textbf{\# Params}} & \multicolumn{2}{c|}{\textbf{Known domains}} & \multicolumn{5}{c}{\textbf{Unknown domains}} \\ \cline{4-10}
& & & \textbf{DRISHTI} & \textbf{REFUGE} & \textbf{BOSCH} & \textbf{FORUS} & \textbf{REMIDIO} & \textbf{ORIGA} & \textbf{RIMONE} \\ \hline
\multirow{3}{*}{\rotatebox{90}{\textbf{OC}}}
& \cite{senbi2025evanyseg} (red channel) & 85,799,425 & 0.1776$^*$ & 0.1483$^*$ & 0.1883$^*$ & 0.1418$^*$ & 0.1594$^*$ & 0.1696$^*$ & \textbf{0.1725} \\ \cline{2-10}
& \cite{senbi2025evanyseg} (added channel) & 85,996,033 & 0.1467$^*$ & 0.1269$^*$ & 0.1855$^*$ & 0.1002$^*$ & 0.1452$^*$ & 0.1683$^*$ & 0.1788 \\ \cline{2-10}
& $Q_{\phi}$ & 11,246,402 & \textbf{0.0669} & \textbf{0.0553} & \textbf{0.0479} & \textbf{0.0466} & \textbf{0.0631} & \textbf{0.0731} & 0.1780 \\ \hline\hline
\multirow{3}{*}{\rotatebox{90}{\textbf{OD}}}
& \cite{senbi2025evanyseg} (red channel) & 85,799,425 & 0.0395$^*$ & 0.0300$^*$ & 0.0625$^*$ & 0.0946$^*$ & 0.0603$^*$ & 0.0584$^*$ & \textbf{0.0430} \\ \cline{2-10}
& \cite{senbi2025evanyseg} (added channel) & 85,996,033 & 0.0272$^*$ & 0.0261$^*$ & 0.0545$^*$ & 0.0681$^*$ & 0.0493 & 0.0388$^*$ & 0.1228$^*$ \\ \cline{2-10}
& $Q_{\phi}$ & 11,246,402 & \textbf{0.0118} & \textbf{0.0188} & \textbf{0.0201} & \textbf{0.0222} & \textbf{0.0402} & \textbf{0.0363} & 0.0473 \\ \hline
\end{tabular}%
}
\end{table*}

Finally, Table~\ref{tab:dice-predictor-results} reports results obtained in quality estimation using our multitask ResNet-based approach and two structure-specific EvanySeg baselines~\cite{senbi2025evanyseg}, one blending the segmentation mask in the red channel of the input image and one concatenating the mask as an additional image channel, both for OC and OD. The proposed Dice predictor achieved the lowest MAE for both OD and OC on all known domains and on all unknown domains except RIMONE, where the best EvanySeg reported a smaller MAE than our method (but without statistical differences). \textcolor{black}{Predicted and actual Dice scores were also strongly correlated across all test images (Pearson $r=0.74$ for OC and $r=0.91$ for OD), exceeding both EvanySeg variants ($r\leq0.61$ for OC, $r\leq0.85$ for OD). Per-dataset $r$ values are included in the results repository.$^1$}

\section{Discussion}

Our work extends the Noisy Student strategy in~\cite{moris2024noisystudent} for domain-generalized OD/OC segmentation by weighting foreground pseudo-labels according to their predicted quality. Because the Teacher, pseudo-labels, architecture, and data splits are unchanged, the gains can be attributed to this weighting strategy. Our method outperforms~\cite{moris2024noisystudent} on the known and most unknown domains, with the largest gains for OC, the harder structure. As shown in Fig.~\ref{fig:qualitative-comparison}, weighting helps limit the propagation of Teacher errors, including false positives from illumination artifacts, background structures, and camera text, while preserving accurate baseline segmentations.

RIMONE is the main exception. Our method improves OC over~\cite{moris2024noisystudent}, but OD remains below~\cite{zhou2022generalizable}, as in the baseline Student. This may be due to RIMONE images being cropped around the optic nerve head (see Fig.~\ref{fig:qualitative-comparison}), which enlarges the apparent OD and changes the spatial context relative to the uncropped training images, in contrast with the OD size and position learned by the Student. This hypothesis requires further testing. Possible remedies include scale-aware augmentations, optic-disc localization, or mixed cropped and uncropped training data. \textcolor{black}{A similar OD gap relative to~\cite{zhou2022generalizable} occurs on ORIGA, where our largest gains are instead for OC. It remains unclear however which factors of the image produce this trade-off between structures. Since \cite{zhou2022generalizable} explicitly addresses appearance-related domain shifts through frequency-domain augmentation and domain-specific image restoration, one potential reason behind this performance differences could be the extreme difference in retinal pigmentation observed in this set with respect to the scans used for training (see Fig.~\ref{fig:qualitative-comparison}). This could also explain why the Teacher in~\cite{moris2024noisystudent} performed better than the Student. Nevertheless, additional experiments and validations are needed to confirm this observation.}

Table~\ref{tab:dice-mean-std} also shows that filtering $\mathcal{D}_U$ by a minimum predicted Dice does not outperform the baseline. Thresholds of $\tau=0.1$ and $\tau=0.3$ reduce the pseudo-labeled set from 16,200 images to 13,000 and 8,453, respectively. The resulting loss of variability may offset the benefit of removing unreliable pseudo-labels. Thus, low-quality foreground pseudo-labels appear more useful when down-weighted than discarded, supporting our choice of foreground weights in $[1,2]$.

The Dice predictor also performed well despite its simple design. The mask-only ResNet outperformed Vision Transformer variants from~\cite{senbi2025evanyseg} in almost all cases, using fewer parameters. Using the fundus image may reintroduce domain shifts linked to image appearance. By contrast, $Q_{\phi}$ shows similar performance on seen and unseen domains in Table~\ref{tab:dice-predictor-results}, together with high correlations. A mask-only input may therefore provide a simpler and more robust signal by focusing on shape, size, and OD/OC spatial relations. However, it may also be influenced by anatomical plausibility, assigning low scores to atypical but correct masks or high scores to well-shaped masks in the wrong location. The low MAE on unknown domains in Table~\ref{tab:dice-predictor-results} suggests a limited effect here, but robustness to atypical morphologies requires further study.

In conclusion, quality-weighted pseudo-labels improve Noisy Student training for domain-generalized OD/OC segmentation without extra annotations or inference-time cost.

\section*{Acknowledgments}
This study was partially funded with a PIP 2021-2023 2472CO (CONICET) and PIP 2025-2027 1273CO (CONICET).

\clearpage
\bibliographystyle{IEEEtran}
\bibliography{mybibliography}

\end{document}